\documentclass[conference]{IEEEtran}
\IEEEoverridecommandlockouts
\usepackage{cite}
\usepackage{amsmath,amssymb,amsfonts,amsthm}
\usepackage{algorithmic}
\usepackage{graphicx}
\usepackage{textcomp}
\usepackage{xcolor}
\usepackage{hyperref}
\usepackage{booktabs, ragged2e}
\usepackage{multirow}
\usepackage[flushleft]{threeparttable}
\usepackage{makecell}

\theoremstyle{definition}
\newtheorem{thm}{Theorem}
\newtheorem{lem}{Lemma}
\newtheorem{asmp}{Assumption}
\newtheorem{prop}{Proposition}

\newtheorem{defn}{Definition}

\makeatletter
\newenvironment{proof*}[1][\proofname]{\par
  \pushQED{\qed}%
  \normalfont \partopsep=\z@skip \topsep=\z@skip
  \trivlist
  \item[\hskip\labelsep
        \itshape
    #1\@addpunct{.}]\ignorespaces
}{%
  \popQED\endtrivlist\@endpefalse
}
\makeatother

\makeatletter
\def\thm@space@setup{\thm@preskip=1pt
\thm@postskip=0pt}
\def\defn@space@setup{\defn@preskip=1pt
\defn@postskip=0pt}
\makeatother

\def\BibTeX{{\rm B\kern-.05em{\sc i\kern-.025em b}\kern-.08em
    T\kern-.1667em\lower.7ex\hbox{E}\kern-.125emX}}

\begin{document}

\title{AnycostFL: Efficient On-Demand Federated Learning over Heterogeneous Edge Devices
}

\author{\IEEEauthorblockN{Peichun Li\IEEEauthorrefmark{1}\textsuperscript{,}\IEEEauthorrefmark{2},
Guoliang Cheng\IEEEauthorrefmark{1}, 
Xumin Huang\IEEEauthorrefmark{1}\textsuperscript{,}\IEEEauthorrefmark{2},
Jiawen Kang\IEEEauthorrefmark{1},
Rong Yu\IEEEauthorrefmark{1},
Yuan Wu\IEEEauthorrefmark{2},
and
Miao Pan\IEEEauthorrefmark{3}
}
\IEEEauthorblockA{\IEEEauthorrefmark{1}School of Automation, Guangdong University of Technology, Guangzhou, China\\
\IEEEauthorrefmark{2}State Key Laboratory of Internet of Things for Smart City, University of Macau, Macau, China\\
\IEEEauthorrefmark{3}Department of Electrical and Computer Engineering, University of Houston, Houston, USA\\
Email: peichun@mail2.gdut.edu.cn, guoliang\_cheng@126.com, huangxu\_min@163.com, \\
\{kavinkang, yurong\}@gdut.edu.cn, yuanwu@um.edu.mo, mpan2@uh.edu}}

\maketitle

\begin{abstract}
In this work, we investigate the challenging problem of on-demand federated learning (FL) over heterogeneous edge devices with diverse resource constraints. We propose a cost-adjustable FL framework, named AnycostFL, that enables diverse edge devices to efficiently perform local updates under a wide range of efficiency constraints. To this end, we design the model shrinking to support local model training with elastic computation cost, and the gradient compression to allow parameter transmission with dynamic communication overhead. An enhanced parameter aggregation is conducted in an element-wise manner to improve the model performance. Focusing on AnycostFL, we further propose an optimization design to minimize the global training loss with personalized latency and energy constraints. 
By revealing the theoretical insights of the convergence analysis, personalized training strategies are deduced for different devices to match their locally available resources. Experiment results indicate that, when compared to the state-of-the-art efficient FL algorithms, our learning framework can reduce up to 1.9 times of the training latency and energy consumption for realizing a reasonable global testing accuracy. Moreover, the results also demonstrate that, our approach significantly improves the converged global accuracy.
\end{abstract}


\section{Introduction}
Federated learning (FL) is an emerging distributed learning paradigm that enables multiple edge devices to train a common global model without sharing individual data \cite{mcmahan2016comm}. This privacy-friendly data analytics technique over massive devices is envisioned as a promising solution to realize pervasive intelligence \cite{qu2021empowering}. However, in many real-world application areas, mobile devices are often equipped with different local resources, which raises the emerging challenges for locally on-demand training \cite{yu2021toward}. Given different local resources status (e.g., computing capability and communication channel state) and personalized efficiency constraints (e.g., latency and energy), it is crucial to customize training strategies for heterogeneous edge devices.

\begin{figure}[t]\centering
  \includegraphics[width=0.45\textwidth]{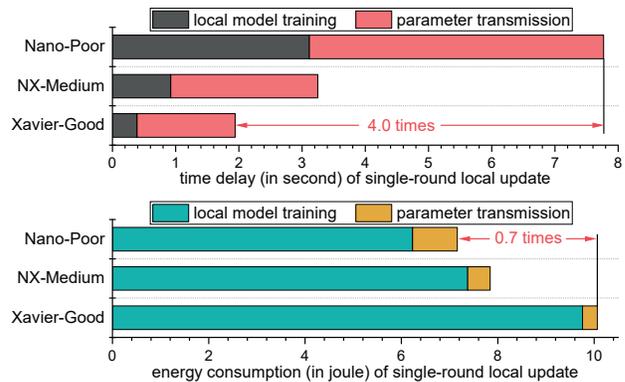}
  \vspace{-7pt}
\caption{The time delay \textbf{(top)} and energy consumption \textbf{(bottom)} of single-round local update on different hardware platforms with varying communication conditions.} \label{fig:pre-exp}
  \vspace{-17pt}
\end{figure}

We perform an in-depth analysis on the time delay and the energy consumption for performing the local model updates at edge devices. 
Specifically, we evaluate and record the cost of local training on three different NVIDIA Jetson family platforms (i.e., Nano, NX AGX, and Xavier AGX) under different channel states (i.e., good, medium, and poor). On the one hand, we observe that the learning efficiency differs significantly  with diverse learning scenarios. As shown in Fig.~\ref{fig:pre-exp}, the single-epoch training on Nano with poor communication condition consumes about 4.0 times training latency than that of Xavier AGX with good communication condition, while its energy consumption is about 0.7 times less than the latter one's. On the other hand, we observe that the bottlenecks of latency and energy are induced by parameter transmission and local model training, respectively.  

The above observations provide insights for a proper design of the on-demand FL system. To handle the resource heterogeneity, it is suggested to alleviate the energy and the latency cost of the local device. More importantly, the computation and communication costs should be jointly reduced to achieve efficient local training. In the literature, most existing studies either employ resource allocation and device scheduling to mitigate the system cost \cite{nishio2019client, chen2020joint, yang2020energy, dinh2020federated, yao2020enhancing, tran2019federated, wu2022non}, or design gradient compression to accelerate the parameter transmission procedure \cite{sattler2019robust, li2021talk, 9685582, 9305988, cui2022optimal, kang2022communication, prakash2021squafl}. The former method inherits the ideas of traditional design for mobile edge systems and takes no account of the optimization for neural networks, while the latter overlooks the computation cost of local model training.

In this paper, we propose ``anycost" FL, named AnycostFL, to break the latency and energy bottlenecks for on-demand distributed training over heterogeneous edge devices. Our goal is to develop a cost-adjustable FL framework that enables edge devices to perform local updates under diverse learning scenarios. To this end, we first design the model shrinking and gradient compression to enable adaptive local updates with different computation and communication costs. Meanwhile, an enhanced parameter aggregation scheme is proposed to fuse the knowledge of the local updates. Following that, we investigate the on-demand learning of AnycostFL by regulating the local model structure, gradient compression policy and computing frequency under personalized latency and energy constraints. However, customizing training strategy for different learning scenarios is a non-trivial task, since how the global accuracy is affected by the local model structure and compression rate is still unknown. To address this issue, we theoretically reveal the convergence insights of our framework, which are further leveraged to guide optimization analysis. Finally, the optimal training strategy is derived for each device according to its locally available resource.

Our main contributions are summarized as follows.
\begin{itemize}
	\item We propose a novel FL framework, named AnycostFL, that enables the local updates with elastic computation cost and communication overhead.
	\item We theoretically present the optimal aggregation scheme and convergence analysis for AnycostFL.
	\item We investigate the on-demand training problem of AnycostFL, and the optimal training strategy is devised to adapt the locally available resource.
	\item Extensive experiments indicate that the proposed AnycostFL outperforms the state-of-the-art efficient FL methods in terms of resource utilization and learning accuracy.
\end{itemize}

The remainder of this paper is organized as follows. Section II describes related studies. In Section III, we detail the main operations of AnycostFL to fulfill the single-round training. 
The problem formulation, theoretical analysis and the corresponding solution are provided in Section IV. 
The experiment evaluations are presented in Section V, and we finally conclude the paper in Section VI and discuss the future directions.

\section{Related Work}
\textit{ Resource Management Methods}. Resource management methods aim to reduce the FL system cost by arranging the local and system resources. Resource allocation methods employ frequency scheduling \cite{9964279}, transmission power control \cite{liu2020privacy}, and bandwidth allocation\cite{xu2021bandwidth} to balance the cost of local training. Recent device selection methods directly exclude those weak devices with poor computation or communication capabilities to accelerate the convergence time \cite{wang2021device, shi2020joint, deng2021auction}. Besides, topology-aware management is another very effective method to mitigate the network throughput \cite{liu2020client, wang2021resource, 9964279}. However, these methods inherit the ideas of the efficient design for traditional mobile systems and overlook the optimization of neural networks.

\textit{ Neuron-aware Techniques}. Neuron-aware techniques focus on revealing the black box of neural networks to improve the training efficiency of the FL system. Early gradient compression utilizes sparsification \cite{sattler2019robust, han2020adaptive}, and quantization\cite{9305988, reisizadeh2020fedpaq, shlezinger2020federated} to reduce the transmission cost of FL system. In addition, feature maps fusion and knowledge distillation can be carried out to improve the information aggregation \cite{lin2020ensemble, zhu2021data}. Besides, FedMask proposes to train a personalized mask for each device to improve the test accuracy on the local dataset \cite{li2021fedmask}. Recently, model structure pruning enables multiple devices with different model architectures to train a shared global model \cite{diao2020heterofl, baek2022joint}. Such methods can reduce the cost of local training, but how to customize optimal training strategies (e.g., gradient compression and model pruning policy) for different learning scenarios is still unknown.

\begin{figure*}[t]\centering
  \includegraphics[width=0.94\textwidth]{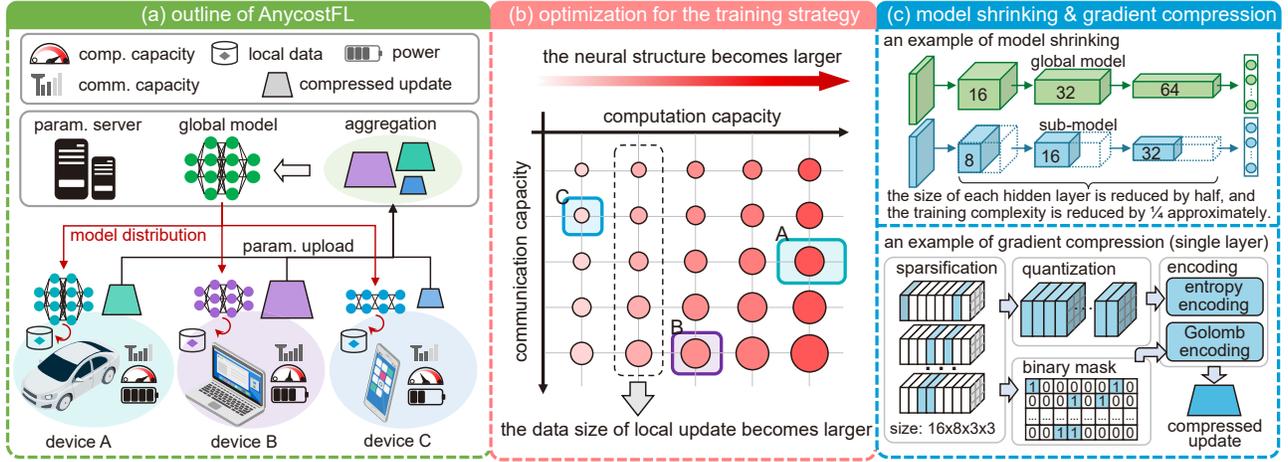}
  \vspace{-4pt}
  \caption{\textbf{left}: AnycostFL over heterogeneous edge devices. \textbf{middle}: the neural structure and gradient compression strategies are customized for diverse devices according to their locally available resources; the darker color indicates the higher computing complexity for training and the larger marker size denotes the larger data size of the local update. \textbf{right}: illustrations of the model shrinking for the local model and the gradient compression for the local update.
}\label{fig:sys-model}
  \vspace{-14pt}
\end{figure*}

\section{Training with AnycostFL}
In this section, we first outline the overall design of AnycostFL. Next, we detail the key techniques of our framework, including elastic model shrinking (EMS), flexible gradient compression (FGC), and all-in-one aggregation (AIO).

\subsection{Outline of AnycostFL}
We consider a generic application scenario of FL with a set of $I$ edge devices ${\cal I} = \{1,2,\cdots,I\}$. We use ${\cal D}_i$ to denote the local training data of the device $i$, and ${\cal D} = { \cup }_{i=1}^I {\cal D}_{i}$ indicates the global data. Let $F_i({\boldsymbol w}) = \ell({\boldsymbol w}, {\cal D}_i)$ represent the local training loss of device $i$ with respect to model weight ${\boldsymbol w}$, where $\ell(\cdot, \cdot)$ is the predetermined loss function. The objective of the FL system is to minimize the following global loss function
\begin{equation}\label{eqn:learn-obj}
 {F({\boldsymbol w}) \buildrel \Delta \over = \sum\limits_{i = 1}^I {\frac{{\left| {\cal D}_i \right|}}{{\left| {\cal D} \right|}}} {F_i}({\boldsymbol w})},
\end{equation} 
where ${\left| {\cal D}_i \right|}$ is the size of ${\cal D}_i$. Given the specified learning task, the original training workload of single sample $W$ and the data size of uncompressed gradient $S$ can be empirically measured.

As shown in Fig.~\ref{fig:sys-model}(a), to reduce the computational complexity of the local model training and the communication cost of gradient update transmission, we propose AnycostFL with two device-side techniques, i.e., model shrinking and gradient compression. 
At the $t$-th global iteration of AnycostFL, the device $i$ is enabled to adjust its training workload and gradient size as $W_{t,i}=\alpha_{t,i}W$ and $S_{t,i}=\beta _{t,i}S$, respectively. Here, $\alpha _{t,i}\in(0,1]$ and $\beta _{t,i} \in (0,1]$ are defined as the model shrinking factor and the gradient compression rate, respectively. The training procedure of AnycostFL is summarized as follows. 
\subsubsection{Elastic local training} At the $t$-th global round, the device $i$ downloads the latest global model ${\boldsymbol w}_{t}$ from the parameter server. With the pre-calculated model shrinking factor $\alpha _{t,i}$, the specialized sub-model ${\boldsymbol w}_{t,i}^{\alpha} = \texttt{shrink}({\boldsymbol w}_t, \alpha_{t,i})$ can be efficiently derived, where function $\texttt{shrink}(\cdot, \cdot)$ indicates the operations for model shrinking. Then, the local training is conducted with sub-model ${\boldsymbol w}_{t,i}^{\alpha}$ and local data ${\cal D}_i$, and the updated local sub-model ${\boldsymbol w}_{t+1,i}^{\alpha}$ is obtained. Furthermore, the local gradient update can be acquired as ${\boldsymbol u}_{t,i}={\boldsymbol w}_{t,i}^{\alpha} - {\boldsymbol w}_{t+1,i}^{\alpha}$.
\subsubsection{Flexible gradient upload} To further reduce the uplink traffic, the local device $i$ is motivated to compress the gradient update ${\boldsymbol u}_{t,i}$ before the parameter transmission. With the given compression rate $\beta _{t,i}$, the compressed gradient update $\tilde{{\boldsymbol u}}_{t,i} = \texttt{cmprs}({\boldsymbol u}_{t,i}, \beta _{t,i})$ is uploaded to the server, where $\texttt{cmprs}(\cdot, \cdot)$ is the function for gradient compression.
\subsubsection{Parameter aggregation} The server collects the compressed local updates $\{\tilde{{\boldsymbol u}}_{t,i}\}_{\forall i}$ with different shrinking factors $\{\alpha _{t,i}\}_{\forall i}$ and compression rates $\{\beta _{t,i}\}_{\forall i}$. After that, the global update is calculated by $\tilde{{\boldsymbol u}}_t=\texttt{aioagg}(\{\tilde{{\boldsymbol u}}_{t,i}\}_{\forall i})$, where $\texttt{aioagg}(\cdot)$ is the server-side all-in-one aggregation. Then, the updated global model is computed as ${\boldsymbol w}_{t+1}={\boldsymbol w}_{t} - \tilde{{\boldsymbol u}}_t$.

After the $T$-round training of the above three-step iterations, the final global model ${\boldsymbol w}_T$ is obtained. Before introducing how to customize the values of $\{\alpha _{t,i}\}_{\forall i}$ and $\{\beta _{t,i}\}_{\forall i}$ in Section IV, we illustrate the details of model shrinking, gradient compression and update aggregation in the rest of this section.

\subsection{Elastic Model Shrinking}
We aim to derive the sub-model ${\boldsymbol w}_{t,i}^{\alpha}$ with training complexity of $\alpha_{t,i}W$ from global model ${\boldsymbol w}_{t}$ by reducing the width of the global model. The shrinking operations work as follows. 
\subsubsection{Server-side channel sorting} To avoid incurring extra memory cost for the edge devices, the server first sorts the channels of the latest global model before the model distribution. Given one layer of the weight of the global model, the server sorts the output channels in the current layer in descending order according to their values of L2 norm, and meanwhile, the input channels of the next layer should be sorted accordingly in the same order to maintain the permutation invariance of the whole model \cite{wang2020federated}. 

\subsubsection{Layer-wise uniform shrinking} Next, the server broadcasts the weight of each layer of the global model in a channel-by-channel manner. Instead of downloading the full global model, each device only receives those important parameters from the global model to assemble the local sub-model. 
Here, we utilize the fixed shrinking ratio for each layer in the same sub-model. 
Empirically, given model shrinking factor $\alpha _{t,i}$, we can reduce the size of the hidden layer by $\sqrt{\alpha _{t,i}}$ to acquire the sub-model. For example, as shown in Fig.~\ref{fig:sys-model}(c), when shrinking a global model with hidden sizes of $\{16, 32, 64\}$ under $\alpha _{t,i}=\frac{1}{4}$, we approximately reduce the size of each hidden layer by half as $\{8, 16, 32\}$ to form the sub-model.

At the beginning of the $t$-th global round, all device initialize their local sub-models ${\{{\boldsymbol w}_{t,i}^{\alpha}\}}_{\forall i}$ by choosing the most important channels from the global model ${\boldsymbol w}_t$. In this way, the training complexity is significantly reduced while maintaining the performance of local sub-models. After that, the local training of device $k$ is conducted with sub-model ${\boldsymbol w}_{t,i}^{\alpha}$, which produces the local gradient ${\boldsymbol u}_{t,i}$ with data size of $\alpha _{t,i}S$.

\subsection{Flexible Gradient Compression}
Given the local update ${\boldsymbol u}_{t,i}$ with the desired compression rate $\beta _{t,i}$, we aim to obtain the compressed update $\tilde{{\boldsymbol u}}_{t,i}$ with data size of $\alpha _{t,i} \beta _{t,i}S$. Let $\rho _{t,i}$ and $L_{t,i}$ denote the sparsity rate and the number of quantization levels, respectively. The gradient compression scheme works as follows.

\subsubsection{Kernel-wise sparsification} Without loss of generality, we take the convolution neural network (CNN) as an example to illustrate the sparsification procedure. We aim to acquire the sparse update $\hat{{\boldsymbol u}}_{t,i}$ from ${\boldsymbol u}_{t,i}$. Let ${\boldsymbol u}_{t,i}[k]$ denote the $k$-th kernel of ${\boldsymbol u}_{t,i}$, and ${\boldsymbol u}_{t,i}=\{{\boldsymbol u}_{t,i}[k]\}_{\forall k}$. 
We measure the importance of each kernel and obtain ${\cal N}=\{\|{\boldsymbol u}_{t,i}[k]\|_2\}_{\forall k}$, where $\|\cdot\|_2$ denotes the L2 norm operation. Next, by selecting the $\lceil \rho _{t,i}K\rceil$-th largest value in $\cal N$ as the threshold $\Pi$, the kernel-wise sparsification is expressed as
\begin{equation}\label{eqn:sparse}
\hat{{\boldsymbol u}}_{t,i}[k]  =
  \begin{cases}
    \bf{0} & \text{if $\|{\boldsymbol u}_{t,i}[k]\|_2 < \Pi$,} \\    {\boldsymbol u}_{t,i}[k] & \text{otherwise.}
  \end{cases}
\end{equation}
Meanwhile, the binary mask of $\hat{{\boldsymbol u}}_{t,i}$ is denoted as ${\boldsymbol m}_{t,i}$.

\subsubsection{Probabilistic quantization} Motivated by the studies in \cite{konevcny2016federated, alistarh2017qsgd}, we aim to obtain the quantized update $\tilde{{\boldsymbol u}}_{t,i}$ with the given sparse $\hat{{\boldsymbol u}}_{t,i}$ and the quantization level $L_{t,i}$. Let $u \in \hat{{\boldsymbol u}}_{t,i}$ be a scalar value.  To begin with, we first calculate the magnitude range of the non-zero elements of $\hat{{\boldsymbol u}}_{t,i}$, denoted as $[u_{\min}, u_{\max}]$, where $u_{\min} = \min \{|u|\}_{\forall u\ne0 }$, and $u_{\max} = \max \{|u|\}_{\forall u\ne0}$. Next, let ${\cal Q}=\{Q_l\}_{l=1}^{L_{t,i}}$ denote the set of quantization points, where $Q_l$ is computed by
\begin{equation}\label{eqn:quant-point}
{Q_l} = \frac{{l\left( {u_{\max } - {u_{\min }}} \right)}}{L_{t,i}} + {u_{\min }}.
\end{equation}
For any $u\in \hat{{\boldsymbol u}}_{t,i}$ and $u\ne0$, we can always find a quantization interval $[Q_l, Q_{l+1}]$ such that $Q_l \le |u| \le Q_{l+1}$, and its corresponding quantized value $\tilde{u}$ is further computed by
\begin{equation}\label{eqn:quant}
\tilde{u}=
  \begin{cases}
    \texttt{sgn}(u) \cdot Q_l & \text{with probability $\frac{{{Q_{l + 1}} - |u|}}{{{Q_{l + 1}} - {Q_l}}}$,} \\
    \texttt{sgn}(u) \cdot Q_{l+1} &  \text{otherwise,} \\
  \end{cases}
\end{equation}
where $\texttt{sgn}(\cdot)$ calculates the sign of the given scalar. Furthermore, the set of the quantization indices of all $ \tilde{u} \in \tilde{{\boldsymbol u}}_{t,i}$ is denoted as ${\cal L}_{t,i}=\{l, Q_l=\tilde{u}\}_{\forall \tilde{u}\ne0}$. Now, $\tilde{{\boldsymbol u}}_{t,i}$ can be represented by a tuple of $\{u_{\min}, u_{\max}, L_{t,i}, {\boldsymbol m}_{t,i}, {\cal L}_{t,i}\}$.

\subsubsection{Lossless encoding} Due to the distribution characteristics of $L_{t,i}$ that smaller indices may occur more frequently, we apply entropy coding to reduce the data size \cite{han2015deep, 9305988}. Besides, the sparse binary matrix ${\boldsymbol m}_{t,i}$ can be compressed by Golomb encoding \cite{sattler2019robust, golomb1966run}.

After determining the compression scheme, we can vary the combinations of $\{\rho_{t,i}, L_{t,i}\}$ and record the corresponding compression rates. Based on the results, we can build a piecewise linear function to predict the compression strategy $\{\rho_{t,i}, L_{t,i}\}$ with the given $\beta _{t,i}$. Notably, this function can be efficiently fitted by the server with a rather small amount of public training data (e.g., 16 samples) in an offline manner.

\subsection{All-in-One Aggregation}
After all the devices upload their encoded updates, the server receives, decodes and then reconstructs the compressed local updates $\{\tilde{{\boldsymbol u}}_{t,i}\}_{\forall i}$. Our goal is to obtain the global update $\tilde{\boldsymbol u}_t$ by aggregating $\{\tilde{{\boldsymbol u}}_{t,i}\}_{\forall i}$. However, the aggregation of local updates in our framework cannot be supported by conventional $\texttt{FedAvg}$ \cite{mcmahan2016comm}, since the local updates are produced by different model structures with different levels of precision (i.e., different quantization levels and sparsity).

\begin{figure}[t]\centering
  \includegraphics[width=0.48\textwidth]{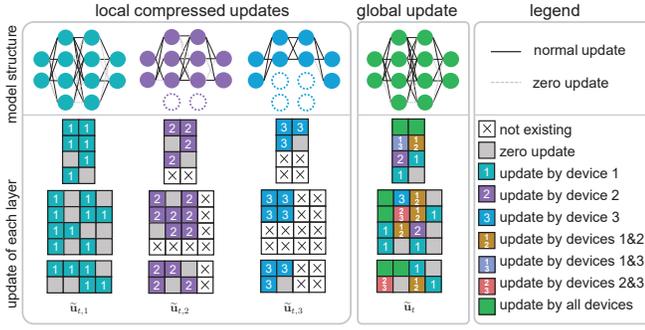}
  \vspace{-5pt}
\caption{An illustration of the all-in-one aggregation.} \label{fig:aioagg}
  \vspace{-15pt}
\end{figure}

To tackle the above challenge, we propose an all-in-one aggregation scheme that fuses the local updates in an element-wise manner. 
Let the set $\{1,2,\cdots,J\}$ index elements of the global update $\tilde{\boldsymbol u}_t$, and $\tilde{\boldsymbol u}_t^{[j]}$ denote the $j$-th element of $\tilde{\boldsymbol u}_t$. 
To accomplish the aggregation for $\tilde{\boldsymbol u}_t^{[j]}$, we first determine the subset of devices ${\cal I}_{j}\subseteq {\cal I}$ whose local model structure also contains the $j$-th element.
Then, we have
\begin{equation}\small\label{eqn:agg}
\tilde{\boldsymbol u}_t^{[j]}=
  \begin{cases}
    0 & \text{if ${\sum\limits_{i \in {\cal I}_{j}} {{{\boldsymbol m}_{t,i}^{[j]}}} }=0$,} \\
    \frac{1}{{\sum\limits_{i \in {\cal I}_{j}} {{p_{t,i}}{{\boldsymbol m}_{t,i}^{[j]}}} }}\sum\limits_{i \in {\cal I}_{j}} {{p_{t,i}}{{\boldsymbol m}_{t,i}^{[j]}}{\boldsymbol u}_{t,i}^{[j]}} &  \text{otherwise,} \\
  \end{cases}
\end{equation}
where $p_{t,i}$ is the aggregation coefficient for the $j$-th device at the $t$-th global round. The optimal values of $\{p_{t,i}\}_{\forall i}$ will be further analyzed in Section IV. 
Fig.~\ref{fig:aioagg} gives an example to illustrate the aggregation details. Specifically,  different elements in the global update are updated by different subsets of devices, and more important elements will ``absorb" knowledge from more devices.
When the $j$-th element is zeroed out by all the devices in ${\cal I}_{j}$, we have $\tilde{\boldsymbol u}_t^{[j]}=0$.

\section{Theoretical Analysis and Optimization}
In this section, we focus on the optimization of our framework by customizing the training strategies for diverse devices. We first formulate the on-demand training problem of AnycostFL. Then, we derive the upper bound of the convergence rate and reveal the key insights to improve the performance of AnycostFL. Based on the analysis, the optimization problem is transformed into a tractable form, and the closed-form solution is derived.

\subsection{AnycostFL over Wireless Networks}
In this subsection, we formulate the computation and communication models for our framework. After that, we build up an on-demand learning problem that minimizes the global training loss with given delay and energy constraints. 

\subsubsection{Computation model}
For the device $i$ at the $t$-th global round, given the model shrinking factor $\alpha _{t,i}$ and computing frequency $f_{t,i}$, the time consumption of local model training can be measured by
\begin{equation}\label{eqn:update-time}
T_{t,i}^{\text{cmp}} = \frac{{\tau |{\cal D}_i|\alpha_{t,i} W}}{f_{t,i}},
\end{equation}
where $\tau$ denotes the number of local epochs. Meanwhile, the corresponding energy consumption can be given by
\begin{equation}\label{eqn:update-energy}
E_{t,i}^{\text{cmp}} = \epsilon _{i} f_{t,i}^2 {\tau |{\cal D}_i|\alpha_{t,i} W},
\end{equation}
where $\epsilon _{i}$ is the hardware energy coefficient of the device $i$.

\subsubsection{Communication model} We consider the frequency division multiple access (FDMA) scheme for the transmission of the local gradient update. 
For the device $i$ at the $t$-th global round, the achievable transmitting rate can be estimated by
\begin{equation}\label{eqn:trans-rate}
r_{t,i} = {b_i}{\log _2}\Big(1 + \frac{{|h_{t,i}|P_{t,i}^{\text{com}}}}{{{N_0}{b_i}}}\Big),
\end{equation}
where $P_{t,i}^{\text{com}}$ is the transmitting power; $b_i$ is the achievable bandwidth; $|h_{t,i}|$ denotes the path loss of wireless channel; $N_0$ is the power spectral density of the additive white Gaussian noise.
For the device $i$ at $t$-th global round, given the update $\tilde{{\boldsymbol u}}_{t,i}$ generated by the local model with a shrinking factor of $\alpha _{t,i}$ and compression rate of $\beta _{t,i}$, the required time $T_{t,i}^{\text{com}}$ and energy consumption $E_{t,i}^{\text{com}}$ of uplink transmission can be  respectively measured by
\begin{equation}\small\label{eqn:trans-time}
T_{t,i}^{\text{com}} = \frac{{{\alpha _{t,i}}{\beta _{t,i}}S}}{{{r_{t,i}}}},~\text{and }
E_{t,i}^{\text{com}} = T_{t,i}^{\text{com}}P_{t,i}^{\text{com}}.
\end{equation}

With the above computation and communication models, we next focus on the optimization problem of AnycostFL. 

\subsubsection{Problem formulation}
To optimize AnycostFL, we study an on-demand training problem. Specifically, the shared maximal latency for each round $T^{\max}$ is determined by the server. The local energy consumption budget for each round $E_{t,i}^{\max}$ is customized by the device itself. Given multiple devices with diverse local resources (e.g., computation, communication and data), our goal is to customize the training strategy for each device to minimize the global training loss with personalized constraints (e.g., latency and energy). To sum up, at the $t$-th global round, we aim to optimize the following problem.
\begin{subequations}\label{eqn:pbm-1}
\begin{align}
    ({\text{P1}}) && \min  F\big({{\boldsymbol w}_t};\{ \alpha &_{t,i} \}_{\forall i} ,\{ \beta_{t,i} \}_{\forall i} \big)    &\tag{\ref{eqn:pbm-1}}\\
    {\text{subject to:}} && T_{t,i}^{\text{cmp}} + T_{t,i}^{\text{com}} &\le {T^{\max }},\forall i,&\label{eqn:p1-ctr-1}\\
    {} && E_{t,i}^{\text{cmp}} + E_{t,i}^{\text{com}} &\le {E_{t,i}^{\max }},\forall i,&\label{eqn:p1-ctr-2}\\
    {}&& {\alpha^{\min}} \le {\alpha _{t,i}} & \le 1,\forall i, & \label{eqn:p1-ctr-3}\\
    {}&& 0 \le {\beta _{t,i}} & \le \beta ^{\max},\forall i, & \label{eqn:p1-ctr-4}\\
    {}&& f_{i}^{\min} \le f_{t,i} &\le f_{i}^{\max},\forall i, &\label{eqn:p1-ctr-5}\\
	{{\textrm{variables:}}}&& \{\alpha _{t,i}, \beta &_{t,i}, f_{t,i}\}_{\forall i},&\nonumber
\end{align}
\end{subequations}
where $F\big({{\boldsymbol w}_t};\{ \alpha _{t,i} \}_{\forall i} ,\{ \beta_{t,i} \}_{\forall i} \big)$ denotes the global loss of the $t$-th round with given the global model weight ${\boldsymbol w}_t$ under the training strategies of $\{ \alpha _{t,i} \}_{\forall i}$ and $\{ \beta_{t,i} \}_{\forall i}$. In the rest of this section, we analyze the relationship between training loss and training strategies. After that, Problem (P1) is further solved based on the theoretical insights.

\subsection{Assumptions and Key Lemmas}
Being in line with the studies in \cite{chen2020joint, chen2020convergence}, we make the following assumptions for the local loss function $F_i, \forall i$.
\begin{asmp}
\label{asmp:1}
$F_{i}$ is $\lambda$-Lipschitz: $\left\Vert {F_{i}}({\boldsymbol w}) - {F_{i}}({\boldsymbol w}^{\prime}) \right\Vert \le \lambda \left\Vert {\boldsymbol w} - {\boldsymbol w}^{\prime} \right\Vert$, where $\lambda>0$.
\end{asmp}
\begin{asmp}
\label{asmp:2}
$F_{i}$ is $\nu$-strongly convex: $F_i({\boldsymbol w}) \ge F_i({\boldsymbol w}^{\prime}) + ({\boldsymbol w} - {\boldsymbol w}^{\prime})^\top \nabla {F_i}({\boldsymbol w^{\prime}}) + \frac{\nu}{2} \left\Vert {\boldsymbol w} - {\boldsymbol w}^{\prime} \right\Vert ^2$.
\end{asmp}
\begin{asmp}
\label{asmp:3}
$F_{i}$ is twice-continuously differentiable. Based on Assumptions \ref{asmp:1} and \ref{asmp:2}, we have $\nu {\bf I} \preceq  \nabla ^2F_i({\boldsymbol w}) \preceq \lambda {\bf I}$.
\end{asmp}
\begin{asmp}
\label{asmp:4}
The ratios between the norms of $ \nabla F_i({\boldsymbol w})$ and  $\nabla F({\boldsymbol w})$ are bounded: $\left\Vert \nabla F_i({\boldsymbol w}) \right\Vert^2 \le \varepsilon \left\Vert \nabla F({\boldsymbol w}) \right\Vert^2$, where $\varepsilon \ge 0$ is a positive constant.
\end{asmp}
\begin{asmp}
\label{asmp:5}
For the moderate shrinking factor $\alpha \ge \alpha ^{\min}$, the first-shrinking-then-training can be approximated as first-training-then-shrinking: $\nabla F_i({\boldsymbol w}^{\alpha})=[\nabla F_i({\boldsymbol w})]^{\alpha}$. Here, we use $[\nabla F_i({\boldsymbol w})]^{\alpha}$ to denote the shrinking operation for $\nabla F_i({\boldsymbol w})$.
\end{asmp}
Next, we give the following two definitions.
\begin{defn}[Local gradient divergence]
The local gradient divergence $\delta _{t,i}$ is defined as the difference between ${\boldsymbol u}_{t,i}$ and $\tilde{{\boldsymbol u}}_{t,i}$, which is given by $\delta _{t,i} = \| {\boldsymbol u}_{t,i} - \tilde{{\boldsymbol u}}_{t,i}\|$.
\label{defn:local-div}
\end{defn}
\begin{defn}[Global gradient divergence]
The global gradient divergence $\Delta _{t}$ is defined as the difference between ${\boldsymbol u}_{t}$ and $\tilde{{\boldsymbol u}}_{t}$, which is measured by
$
\Delta _{t} = \| {\boldsymbol u}_{t} - \tilde{{\boldsymbol u}}_{t}\| = {\Big\| {\sum\limits_{i = 1}^I {{p_{t,i}} } {{\boldsymbol u}}_{t,i}} -  {\sum\limits_{i = 1}^I {{p_{t,i}} } \tilde{{\boldsymbol u}}_{t,i}} \Big\|}$.
\label{defn:global-div}
\end{defn}
Notably, in Definition \ref{defn:local-div}, ${\boldsymbol u}_{t,i}$ and $\tilde{{\boldsymbol u}}_{t,i}$ may have different dimensions. We pad the missing elements in $\tilde{{\boldsymbol u}}_{t,i}$ with zeros before the arithmetic operation.
Next, we are interested in how the training strategies $\{ \alpha _{t,i}, \beta_{t,i} \}_{\forall i}$ affect $\{\delta _{t,i}\}_{\forall i}$ and $\Delta _{t}$. We derive the following two lemmas.

\begin{lem}
For the local training with the model shrinking factor $\alpha _{t,i}$ and compression rate $\beta _{t,i}$. The square of the local gradient divergence is bounded by 
\vspace{-5pt}
\begin{equation}\label{eqn:loc-divergence}
{\mathbb E} \|\delta _{t,i}\|^2 \le\big(1 - \alpha _{t,i} (2-\alpha _{t,i})\sqrt {\beta _{t,i}}  \big)^2{\mathbb E} \| {\boldsymbol u}_{t,i} \|^2.
\end{equation}
\label{lem:local-divergence}
\vspace{-5pt}
\end{lem}
\vspace{-5pt}
\begin{proof*}
See Appendix A.
\end{proof*}
\begin{lem}
For the local update $\{\tilde{{\boldsymbol u}}_{t,i}, \forall i\}$ with the corresponding training strategies $\{ \alpha _{t,i}, \beta_{t,i} \}_{\forall i}$ and aggregation coefficients $\{p_{t,i}\}_{\forall i}$, the square of the global gradient divergence is bounded by 
\vspace{-10pt}
\begin{align}
\small
\begin{split}
\label{eqn:glb-divergence}
{\mathbb E} {\left\Vert \Delta _t  \right\Vert^2} \le I\varepsilon \eta ^2 { {\sum\limits_{i = 1}^I {p_{t,i}^2}\big(1 - \alpha _{t,i} (2-\alpha _{t,i})\sqrt {\beta _{t,i}}  \big)^2} } {\mathbb E}  \|\nabla F({{\boldsymbol w}_t})\|^2.
\end{split}
\end{align}
\label{lem:glb-divergence}
\vspace{-13pt}
\end{lem}
\begin{proof*}
See Appendix B.
\end{proof*}

\subsection{Optimal Aggregation Scheme and Convergence Analysis}
Intuitively, the local update ${\boldsymbol u}_{t,i}$ generated with larger $\{\alpha _{t,i}, \beta_{t,i}\}$ may carry more accurate information, and thus a larger $p_{t,i}$ should be assigned during the aggregation. Based on Lemma \ref{lem:glb-divergence}, we deduce the following theorem.

\begin{thm}[Optimal aggregation scheme]
Given the local updates $\{\tilde{{\boldsymbol u}}_{t,i}\}_{\forall i}$ with corresponding training strategies $\{ \alpha _{t,i}, \beta _{t,i} \}_{\forall i}$, the optimal aggregation coefficients are 
\begin{align}
\begin{split}
p_{t,i}^ \ast  = \frac{{\frac{1}{{{{\big(1 - \alpha _{t,i} (2-\alpha _{t,i})\sqrt {\beta _{t,i}}  \big)}^2}}}}}{{\sum\nolimits_i {\frac{1}{{{{\big(1 - \alpha _{t,i} (2-\alpha _{t,i})\sqrt {\beta _{t,i}}  \big)}^2}}}} }}, \forall i.
\end{split}
\label{eq:thm1}
\end{align}
\label{thm:1}
\end{thm}
\begin{proof*}
Based on Lemma \ref{lem:glb-divergence}, we study the following optimization problem to minimize the global gradient divergence.
\begin{subequations}\label{eqn:pbm-2}\small
\begin{align}
    ({\text{P2}}) && \mathop {\min }\limits_{\{p _{t,i}\}_{\forall i}} \sum\limits_{i = 1}^I {p_{t,i}^2}\big(1 - \alpha &_{t,i} (2-\alpha _{t,i})\sqrt {\beta _{t,i}}  \big)^2    &\tag{\ref{eqn:pbm-2}}\\
    {\text{subject to:}} && p_{t,i} \ge& 0,\forall i, & \label{eqn:p2-ctr-1}\\
    {}&& \sum\limits_{i = 1}^I &{{p_{t,i}}}  = 1. &\label{eqn:p2-ctr-2}
\end{align}
\end{subequations}

It can be verified that Problem (P2) is a convex optimization problem. We further solve the problem by the Karush–Kuhn–Tucker (KKT) conditions. Let $\{\varpi\}_{\forall i}$ and $\theta$ be the Lagrange multipliers for Constraints (\ref{eqn:p2-ctr-1}) and (\ref{eqn:p2-ctr-2}), respectively. Then, we obtain
\vspace{-10pt}
\begin{align}\label{eqn:p2-kkt-cdt}
\small
\begin{split}
\varpi _i \ge 0, \;\; \varpi _i p_{t,i}=0, \;\;p_{t,i} \ge 0,\;\; \sum\limits_{i = 1}^I {{p_{t,i}}}  = 1,\\
2{p_{t,i}} \big(1 - \alpha _{t,i} (2-\alpha _{t,i})\sqrt {\beta _{t,i}}  \big)^2 -\varpi _i +\theta =0, \;\forall i.
\end{split}
\end{align}
Being in line with the study in \cite{9425020}, we can obtain
\begin{equation} \label{eqn:p2-kkt-4}
p_{t,i} = -\frac{\theta}{2 \big(1 - \alpha _{t,i} (2-\alpha _{t,i})\sqrt {\beta _{t,i}}  \big)^2}.
\end{equation}
By putting Eqn.~(\ref{eqn:p2-kkt-4}) into Eqn.~(\ref{eqn:p2-ctr-2}), we obtain
\begin{equation} \label{eqn:p2-kkt-5}
\theta = -\frac{2}{\sum\nolimits_{k} {\frac{1}{\big(1 - \alpha _{t,i} (2-\alpha _{t,i})\sqrt {\beta _{t,i}}  \big)^2}} }.
\end{equation}
Putting Eqn.~(\ref{eqn:p2-kkt-5}) into Eqn.~(\ref{eqn:p2-kkt-4}) completes the proof.
\end{proof*}

With the optimal aggregation scheme, we investigate the upper bound of the convergence rate of AnycostFL.  
\begin{defn}[Local and global learning gains]
The local and global learning gains are defined as $g_{t,i}=\alpha_{t,i}^4\beta_{t,i}$ and $g_t = {\sum\nolimits_{i} {g_{t,i}}} / I$, respectively. Specifically, the local and global learning gains (i.e., $g_{t,i}\in[0,1]$ and $g_t\in[0,1]$) measure the amount of effective information carried in the local and global updates, respectively. 
\label{defn:learning-gain}
\end{defn}

\begin{thm}[Convergence rate of AnycostFL]
Let $g^{\min}=\min \{g_t\}_{\forall t}$ be the minimal global learning gain over the $T$-round training. The upper bound of the convergence rate of AnycostFL satisfies
\begin{align}
\begin{split}\label{eqn:thm-2-0}
{\mathbb E} \big(F({{\boldsymbol w}_{T}}) - F({\boldsymbol w}^\ast)\big) &\le Z^{T-1} {\mathbb E} \big(F({{\boldsymbol w}_{0}}) - F({\boldsymbol w}^\ast)\big),
\end{split}
\end{align}
where $Z=1 - \frac{\nu }{\lambda }\left( {1 - {\varepsilon}( {1 - g^{\min} } )} \right)$.  Recall that parameters $\nu,\lambda$ and $\epsilon$ are defined in Assumptions \ref{asmp:1} to \ref{asmp:4} before. 
\label{thm:2}
\end{thm}
\begin{proof*}
See Appendix C.
\end{proof*}
Based on Definition \ref{defn:learning-gain} and Theorem \ref{thm:2}, we derive the following proposition.
\begin{prop} The key to minimizing the training loss of AnycostFL is to maximize the learning gain $g_t$ for each global round. If $g_t=1~ \forall t$, AnycostFL degrades to conventional FL without model shrinking and gradient compression. 
\label{prop:1}
\end{prop}

\subsection{Solution for Problem (P1)}
Based on Theorem \ref{thm:2} and Proposition \ref{prop:1}, Problem (P1) can be transformed into the following problem.
\begin{subequations}\label{eqn:pbm-3}
\begin{align}
    ({\text{P3}}) && \max ~  \frac{1}{I}&\sum\limits_{t = 1}^I {\alpha_{t,i}^4\beta_{t,i}}     &\tag{\ref{eqn:pbm-3}}\\
    {\text{subject to:}} && \text{Constrains~}&\text{(\ref{eqn:p1-ctr-1}) to (\ref{eqn:p1-ctr-5})},&\nonumber\\
	{{\textrm{variables:}}}&& \{\alpha _{t,i}, \beta &_{t,i}, f_{t,i}\}_{\forall i}.&\nonumber
\end{align}
\end{subequations}

Based on Constraints (\ref{eqn:p1-ctr-1}) and (\ref{eqn:p1-ctr-2}) for the training latency and energy, we obtain the following lemma.
\begin{lem}
The equality will always hold for Constraints (\ref{eqn:p1-ctr-1}) and (\ref{eqn:p1-ctr-2}) when confirming the optimal training strategy $\{\alpha _{t,i}^\ast, \beta _{t,i}^\ast, f_{t,i}^\ast\}_{\forall i}$, and thus $T_{t,i}^\ast=T^{\max}$ and $E_{t,i}=E_{t,i}^\ast~\forall i$.
\label{lem:lem-3}
\end{lem}
\begin{proof*}
The lemma can be proved by showing the contradiction. Suppose that there exists $i_0$ such that $T_{t,i_0}^\ast < T^{\max}$. We can find a new solution $\{\alpha_{t,i_0}^{\prime}, \beta_{t,i_0}^{\ast}, f_{t,i_0}^\prime\}$ for device $i_0$ and $\alpha_{t,i_0}^{\prime} > \alpha_{t,i_0}^{\ast}$, $f_{t,i_0}^\prime < f_{t,i_0}^\ast$, such that $T_{t,i_0}^{\prime} = T^{\max}$ and $E_{t,t_0}^{\prime}=E_{t,i}^{\max}$. Since the global learning gain increases with the increase of $\alpha_{t,i_0}$, we have $g_t^{\prime} > g_t^{\ast}$. Likewise, the contradiction also appears when $E_{t,i_0}^\ast < E_{t,i_0}^{\max}$, and thus we complete the proof.
\end{proof*}

Based on Lemma \ref{lem:lem-3}, we employ two intermediate variables (i.e., $\phi _{t,i}$ and $\varphi _{t,i} $) for each device to reparameterize Problem (P3). Specifically, $\phi_{t,i} \in [0,1]$ and $\varphi_{t,i} \in [0,1]$ are the splitting factors for latency and energy, respectively, such that 
\begin{align}
\small
\begin{split}
\label{eqn:p3-splitting}
T^{\text{cmp}}_{t,i}=\phi_{t,i}T^{\max},\;\; T^{\text{com}}_{t,i}=(1-\phi_{t,i})T^{\max},\\
E^{\text{cmp}}_{t,i}=\varphi_{t,i}E^{\max}_{t,i},\;\; E^{\text{com}}_{t,i}=(1-\varphi_{t,i})E^{\max}_{t,i}, ~\forall i.
\end{split}
\end{align}
By combining Eqns (\ref{eqn:update-time}) and (\ref{eqn:p3-splitting}), the local learning gain of the device $i$ at the $t$-th round can be rewritten as
\begin{align}
\small
\begin{split}\label{eqn:local-gain}
g_{t,i}(\phi_{t,i}) = \kappa _{t,i}\big( E^{\max}_{t,i} - (1 - \phi_{t,i} )T^{\max}P_{t,i}^{\text{com}} \big)(\phi_{t,i} ^2 - \phi_{t,i} ^3 ),
\end{split}
\end{align}
where $\kappa _{t,i} = \frac{{{r_{t,i}}}}{{S{\epsilon _i}}}\big( \frac{{T^{\max }}}{{\tau | {{\cal D}_i} |W}} \big)^3$. 

Note that Problem (P3) can be transformed into $I$ sub-problems because the decision-making procedure of each device is independent. Based on Eqn.~(\ref{eqn:local-gain}), the $i$-th sub-problem can be expressed as a single-variable optimization problem with respect to $\phi_{t,i}$ as follows.
\begin{subequations}\label{eqn:pbm-4}
\begin{align}
    ({\text{P4}}) &&  \mathop {\max }\limits_{\phi_{t,i}}  ~ g_{t,i}\big(\phi_{t,i}&\big)     &\tag{\ref{eqn:pbm-4}}\\
    {\text{subject to:}} && \phi_{t,i}^{\min} \le\phi_{t,i}\le &\phi_{t,i}^{\max},&\nonumber
\end{align}
\end{subequations}
where the lower and upper limits of $\phi_{t,i}$ can be acquired by
\begin{align}
\small
\begin{split}
\phi_{t,i}^{\min} = {\max \Big\{ \frac{{{\alpha ^{\min }}\tau \left| {{{\cal D}_i}} \right|W}}{{f_{i}^{\max }{T^{\max }}}},1 - \frac{{{\beta ^{\max }}S}}{{{r_{t,i}}{T^{\max }}}}\Big\} },\\
\phi_{t,i}^{\max} = \min \Big\{ \frac{{\tau \left| {{{\cal D}_i}} \right|W}}{{f_i^{\min }{T^{\max }}}},1 - \frac{{{\alpha ^{\min }}{\beta ^{\min }}S}}{{{r_{t,i}}{T^{\max }}}}\Big\}.
\end{split}
\end{align}
Based on the first-order optimality condition $\partial g_{t,i}/\phi_{t,i}=0$, we obtain the stationary points as
\begin{equation}
\small
\phi_{t,i}^{\text{s1}}  = \frac{{\sqrt {\psi_{t,i}}  - 3E^{\max}_{t,i}}}{{8P_{t,i}^{\text{com}}T^{\max}}}+\frac{3}{4},~
\phi_{t,i}^{\text{s2}}   =  - \frac{{\sqrt {\psi_{t,i}} +3E^{\max}_{t,i}}}{{8P_{t,i}^{\text{com}}T^{\max}}}-\frac{3}{4},
\end{equation}
where $\psi_{t,i} ={4(P_{t,i}^{\text{com}}{T^{\max}})^2 - 4E^{\max}_{t,i}P_{t,i}^{\text{com}}T^{\max} + 9(E^{\max}_{t,i})^2}$. Let ${\cal S}_{t,i} = \{\phi_{t,i}^{\min}, \phi_{t,i}^{\max}, \phi_{t,i}^{\text{s1}},\phi_{t,i}^{\text{s2}}\}$ denote the union of the stationary points and the boundary points for Problem (P4). Then, ${\cal S}_{t,i}^{\prime} = \{\phi_{t,i}| \phi_{t,i}\in [\phi_{t,i}^{\min}, \phi_{t,i}^{\max}], \phi_{t,i}\in {\cal S}_{t,i}\}$ is the set of the feasible solutions of ${\cal S}_{t,i}$.
The optimal solution for Problem (P4) can be acquired by 
\begin{equation}
\phi_{t,i}^{\ast} = \mathop {\arg \max }\limits_{\phi_{t,i} \in {\cal S}_{t,i}^{\prime}} ~g_{t,i}(\phi_{t,i}).
\end{equation}
Furthermore, we obtain the optimal solution for device $i$ at the $t$-th global round by putting $\phi_{t,i}^{\ast}$ into the following equations.
\begin{align}
\small
\begin{split}
{\varphi_{t,i} ^\ast} = 1 - \frac{{(1 - {\phi_{t,i}^{\ast}}){T^{\max }}P_{t,i}^{{\text{com}}}}}{{E_{t,i}^{\max }}}&,
{\alpha_{t,i} ^\ast} = \sqrt[3]{{\frac{{{{(\phi_{t,i}^{\ast} {T^{\max }})}^2}{\varphi_{t,i} ^\ast} E_{t,i}^{\max }}}{{{\epsilon _i}({{\tau \left| {{{\cal D}_i}} \right|W}})^3}}}},\\
{\beta_{t,i} ^\ast} = \frac{{{r_{t,i}}(1 - {\phi_{t,i}^{\ast}}){T^{\max }}}}{{{\alpha _{t,i} ^\ast}S}}&, {f_{t,i}^{\ast}} = \frac{{{\alpha _{t,i} ^\ast}\tau \left| {{{\cal D}_i}} \right|W}}{{{\phi_{t,i}^{\ast}}{T^{\max }}}}.
\end{split}
\end{align}
\begin{figure*}\centering
  \includegraphics[width=0.98\textwidth]{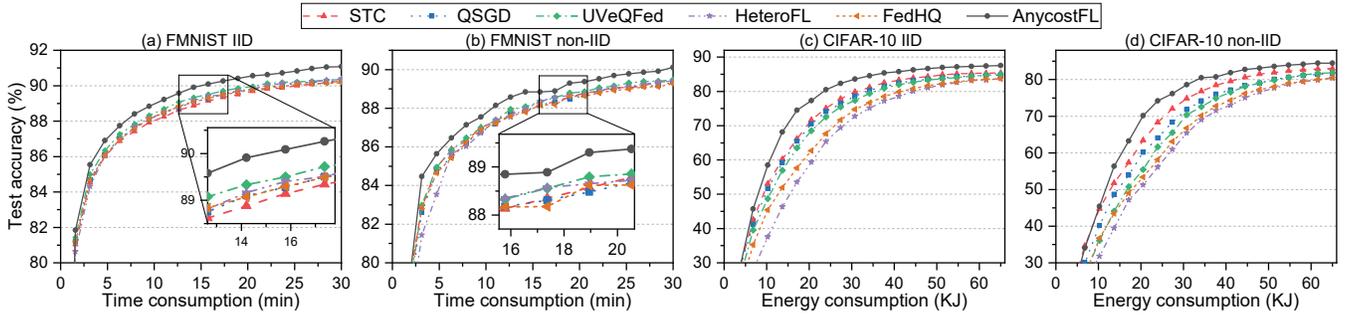}
  \vspace{-8pt}
  \caption{Performance on various network architectures and datasets. ((a-b): global accuracy vs. time consumption with Fashion MNIST on 2-layer CNN; (c-d): global accuracy vs. energy consumption with CIFAR-10 on VGG-9.)  
}\label{fig:compare-acc}
  \vspace{-10pt}
\end{figure*}

\begin{table*}
  \caption{Performance comparison between AnycostFL and other methods on Fashion-MNIST and CIFAR-10 datasets.}
  \label{tab:compare-exp}
  \centering
  \scriptsize
  \setlength{\tabcolsep}{1.6pt}
  \begin{tabular}{llcccccccccccc}
    \toprule
         & & \multicolumn{6}{c}{\textbf{IID} }&\multicolumn{6}{c}{\textbf{non-IID}}\\
    \cmidrule(r){3-8}\cmidrule(r){9-14}
         Dataset & Method & \#Round & \makecell{Energy\\(KJ)} & \makecell{Latency \\(min)} & \makecell{Comp.\\(TFLOPs)} & \makecell{Comm.\\(GB)} & \makecell{Best Acc.\\(\%)} & \#Round & \makecell{Energy\\(KJ)} & \makecell{Latency \\(min)} & \makecell{Comp.\\(TFLOPs)} & \makecell{Comm.\\(GB)} & \makecell{Best Acc.\\(\%)}\\
    \midrule
\multirow{6}{*}{\makecell{\textbf{FMNIST} \\$\{90\%, 89\%\}^\ast$}}
& STC	& 305 (1.7$\times$) 	& 10.94 (1.4$\times$) 	& 25.42 (1.7$\times$) 	& 152.71 	& 0.71  	& 90.28$\pm$0.18
& 283 (1.3$\times$) 	& 10.17 (1.1$\times$) 	& 23.56 (1.3$\times$) 	& 141.53 	& 0.66  	& 89.47$\pm$0.16\\
& QSGD	& 283 (1.6$\times$) 	& 11.40 (1.4$\times$) 	& 23.56 (1.6$\times$) 	& 141.53 	& 0.80  	& 90.39$\pm$0.04
& 279 (1.3$\times$) 	& 11.27 (1.2$\times$) 	& 23.28 (1.3$\times$) 	& 139.86 	& 0.79  	& 89.49$\pm$0.07\\
& UVeQFed & 247 (1.4$\times$) 	& 11.36 (1.4$\times$) 	& 20.58 (1.4$\times$) 	& 123.67 	& 0.72  	& 90.44$\pm$0.10
& 266 (1.2$\times$) 	& 12.21 (1.3$\times$) 	& 22.14 (1.2$\times$) 	& 133.01 	& 0.77  	& 89.64$\pm$0.16\\
& HeteroFL 	&  233 (1.3$\times$) 	& 12.03 (1.5$\times$) 	& 21.78 (1.5$\times$) 	& 92.21 	& 0.57  	& 90.43$\pm$0.13
& 242 (1.1$\times$) 	& 12.51 (1.3$\times$) 	& 22.62 (1.3$\times$) 	& 95.77 	& 0.59  	& 89.42$\pm$0.10 \\
& FedHQ 	& 288 (1.6$\times$) 	& 13.89 (1.7$\times$) 	& 24.03 (1.6$\times$) 	& 144.36 	& 0.86  	& 90.21$\pm$0.07
& 313 (1.5$\times$) 	& 14.96 (1.6$\times$) 	& 26.06 (1.5$\times$) 	& 156.55 	& 0.93  	& 89.27$\pm$0.19 \\
& AnycostFL	& \textbf{179} (1.0$\times$) 	& \textbf{8.07} (1.0$\times$) 	& \textbf{14.94} (1.0$\times$) 	& \textbf{67.49} 	& \textbf{0.35}  	& \textbf{91.20}$\pm$0.09
& \textbf{214} (1.0$\times$) 	& \textbf{9.63} (1.0$\times$) 	& \textbf{17.83} (1.0$\times$) 	& \textbf{80.51} 	& \textbf{0.42}  	& \textbf{90.32}$\pm$0.14 \\
\midrule
\multirow{6}{*}{\makecell{\textbf{CIFAR-10} \\$\{82\%, 80\%\}^\ast$}}
& STC	& 341 (1.2$\times$) 	& 35.39 (1.3$\times$) 	& 56.83 (1.2$\times$) 	& 4160.56 	& 1.78  	& 85.38$\pm$0.29
& 412 (1.1$\times$) 	& 42.39 (1.3$\times$) 	& 68.67 (1.1$\times$) 	& 5026.84 	& 2.15  	& 83.09$\pm$0.53\\
& QSGD	& 337 (1.2$\times$) 	& 39.82 (1.5$\times$) 	& 56.17 (1.1$\times$) 	& 4111.76 	& 2.14  	& 84.83$\pm$0.54
& 430 (1.2$\times$) 	& 50.29 (1.5$\times$) 	& 71.61 (1.2$\times$) 	& 5242.39 	& 2.73  	& 81.94$\pm$0.13\\
& UVeQFed	& 296 (1.0$\times$) 	& 40.77 (1.5$\times$) 	& 49.28 (1.0$\times$) 	& 3607.45 	& 2.12  	& 85.09$\pm$0.16
& 377 (1.0$\times$) 	& 51.59 (1.5$\times$) 	& 62.89 (1.0$\times$) 	& 4603.87 	& 2.71  	& 82.30$\pm$0.28\\
& HeteroFL	&  332 (1.1$\times$) 	& 50.07 (1.9$\times$) 	& 69.14 (1.4$\times$) 	& 3222.26 	& 1.65  	& 83.75$\pm$0.55
& 413 (1.1$\times$) 	& 62.88 (1.9$\times$) 	& 85.78 (1.4$\times$) 	& 3990.49 	& 2.05  	& 80.68$\pm$0.45 \\
& FedHQ	& 340 (1.2$\times$) 	& 48.95 (1.9$\times$) 	& 56.67 (1.2$\times$) 	& 4148.36 	& 2.32  	& 84.02$\pm$0.22
& 435 (1.2$\times$) 	& 61.99 (1.9$\times$) 	& 72.44 (1.2$\times$) 	& 5303.40 	& 2.96  	& 81.00$\pm$0.41 \\
& AnycostFL	& \textbf{294} (1.0$\times$) 	& \textbf{26.43} (1.0$\times$) 	& \textbf{48.94} (1.0$\times$) 	& \textbf{2459.92} 	& \textbf{1.56}  	& \textbf{87.72}$\pm$0.23
& \textbf{372} (1.0$\times$) 	& \textbf{33.51} (1.0$\times$) 	& \textbf{62.06} (1.0$\times$) 	& \textbf{3118.60} 	& \textbf{1.98}  	& \textbf{84.91}$\pm$0.51 \\
    \bottomrule
\multicolumn{10}{l}{\textsuperscript{*}\footnotesize{$\{x,y\}$: $x$ and $y$ denote the target global model accuracy under IID and non-IID data settings, respectively.}}
  \end{tabular}
 \vspace{-6pt}
\end{table*}
\vspace{-8pt}
Notably, the decision-making process of each device does not involve the auxiliary information of the resource status from other devices. At the beginning of each global round, each device can determine its training strategy locally. 

\section{Experiment Evaluations}
\subsection{Experiment Settings}
\subsubsection{Setup for FL training} We consider the FL application with image classification on Fashion-MNIST and CIFAR-10 datasets \cite{xiao2017fashion, krizhevsky2009learning}. For Fashion-MNIST, we use a small convolutional neural network (CNN) with data size of model update as 53.22Mb \cite{mcmahan2016comm}. For the CIFAR-10 dataset, we employ VGG-9 with data size of model update as 111.7Mb \cite{simonyan2014very}. For IID and non-IID data settings, we follow the dataset partition strategy in \cite{wang2020federated}. For the learning hyper-parameters, the learning rate, batch size and local epoch are set as \{0.01, 32, 1\} for Fashion-MNIST and \{0.08, 64, 1\} for CIFAR-10 dataset. The maximal latency is set as $T^{\max}=10$ seconds and the energy budget is set as $E^{\max}_{t,i}\sim U[3, 9]$ joules for the CIFAR-10 dataset, and the corresponding hyper-parameters for the FMNIST dataset are halved by default. Additionally, we set $\alpha^{\min}=1/4$ and $\beta^{\max}=1/15$. 
\subsubsection{Setup for mobile system}
We investigate a mobile system with $I=60$ devices located within a circle cell with a radius of 550 meters, and a base station is situated at the center. To simulate the mobility, the position of each device is refreshed randomly at the beginning of each round \cite{9207871}. 
For the computation, the energy coefficient is set as $\epsilon_i \sim U[5\times10^{-27}, 1\times 1^{-26}$].
For communication, the bandwidth is set as $1$MHz equally for each device, and the path loss exponent is 3.76. The transmission power is set as $0.1$W, and $N_0$ is set as $-114$dBm/MHz.

\subsection{Performance Comparisons}
We compare the proposed AnycostFL with the following efficient FL algorithms with three different random seeds.
\begin{itemize}
	\item {\bf STC}. The sparse ternary compression (STC) is adapted to reduce the cost of uplink parameter transmission \cite{sattler2019robust}.
	\item {\bf QSGD}. The TopK sparsification and probabilistic quantization are combined to compress the local gradient \cite{alistarh2017qsgd}.  
	\item {\bf UVeQFed}. The TopK sparsification and universal vector quantization are used to compress the local gradient \cite{9305988}.
	\item {\bf HeteroFL}. Each device trains the local sub-model in different widths to match its computation capacity \cite{diao2020heterofl}.
	\item {\bf FedHQ}. Each device uses different quantization levels to compress the gradient according to its channel state \cite{9425020}.
\end{itemize}

\begin{figure*}\centering
  \includegraphics[width=0.98\textwidth]{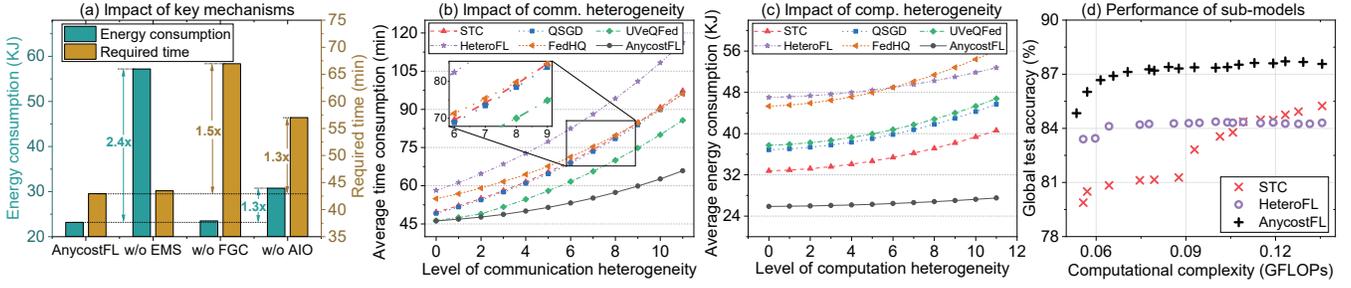}
  \vspace{-4pt}
  \caption{The main advantages of AnycostFL. ((a): the impact of key mechanisms; (b-c): the impact of system heterogeneity; (d): the performance of sub-models.) 
}\label{fig:ablations}
  \vspace{-12pt}
\end{figure*}

Fig.~\ref{fig:compare-acc} shows the performance of the global model over time consumption and energy consumption under the IID and the non-IID data setting. With the same training efficiency (i.e., time and energy consumption), the proposed AnycostFL consistently outperforms the baseline schemes to improve the test accuracy of the global model. Meanwhile, Table \ref{tab:compare-exp} provides the best accuracy and required system cost for achieving the specified test accuracy. Particularly, when compared with HeterFL and FedHQ, AnycostFL can reduce up to 1.9 times the energy consumption to reach the test accuracy of 82\% on CIFAR-10 dataset under the IID setting. When compared with STC, AnycostFL can reduce up to 1.7 times the time consumption to reach the test accuracy of 90\% on FMNIST dataset under the IID setting. Moreover, our framework can significantly improve the best accuracy of the global model by 2.33\% and 1.82\% on CIFAR-10 dataset under the IID and the non-IID settings, respectively.

\subsection{Impact of Key Mechanisms and Hyper-parameters}
Fig.~\ref{fig:ablations}(a) verifies the advantages of the main techniques of AnycostFL. We gradually remove the elastic model shrinking (w/o EMS), the flexible gradient compression (w/o FGC) and the all-in-one aggregation (w/o AIO), and record the required system cost to achieve 80\% test accuracy with CIFAR-10 dataset under the IID setting. We observe that the proposed EMS and FGC can significantly save the energy consumption and training time, respectively. Besides, AIO contributes to saving both energy and time.

We next evaluate the impact of resource heterogeneity on the training efficiency in Fig.~\ref{fig:ablations}(b-c). We set the average energy coefficient $\epsilon_i$ as $7.5\times10^{-27}$ and the average distance between the base station and edge devices as 400 meters, and then change their variances to simulate the computation and communication heterogeneity, respectively. The larger variance indicates a higher level of system heterogeneity. As we expect, the proposed AnycostFL shows more resilience than other baselines to tackle the high level of system heterogeneity.

We also evaluate the performance of sub-models in different widths in Fig.~\ref{fig:ablations}(d). Specifically, We compare AnycostFL with HeteroFL (i.e., local training with different widths) and STC (i.e., the best-performing compression-only method). The sub-models are derived from the well-trained global model without further re-training. Surprisingly, the sub-models of the global model trained by AnycostFL can still maintain satisfactory test accuracy, which provides dynamic inference for diverse edge devices after the training time.

\section{Conclusion}
In this paper, we proposed AnycostFL, a joint computation and communication efficient framework for FL, that enables edge devices with diverse resources to train a shared global model. 
We aimed to minimize the global training loss under given personalized latency and energy constraints.
By leveraging the theoretical insight of AnycostFL, we decomposed the optimization problem into multiple sub-problems. Following that, the optimal training strategy is derived for each device according to its locally available resource.
Experiments demonstrate the advantage of our framework in improving the system efficiency and model performance compared to the state-of-the-art methods. 

\section*{Acknowledgment}
Rong Yu and Yuan Wu are the corresponding authors. This work was supported in part by National Key R\&D Program of China under Grant 2020YFB1807802, in part by National Natural Science Foundation of China under Grants 61971148, 62102099, U22A2054 and 62001125, in part by Science and Technology Development Fund of Macau SAR under Grant 0162/2019/A3, in part by FDCT-MOST Joint Project under Grant 0066/2019/AMJ, in part by the Guangdong Basic and Applied Basic Research Foundation (2022A1515011287), and in part by US National Science Foundation under grant CNS-2107057.

\appendices
\section{Proof of Lemma 1}
\begin{proof*}
For the given local gradient $\tilde{{\boldsymbol u}}_{t,i}$ with shrinking factor $\alpha _{t,i}$ and gradient compression rate $\beta _{t,i}$, we aim to capture the divergence between $\tilde{{\boldsymbol u}}_{t,i}$ and ${\boldsymbol u}_{t,i}$.
Suppose that the absolute value of the element in ${{\boldsymbol u}}_{t,i}$ follows uniform distribution $|u|\sim U(0, u_{\max})$, and $u_{\max} = \max\{|u|\}_{\forall u \in {{\boldsymbol u}}_{t,i}}$.

For clear notation, we sort the element-wise absolute value of ${{\boldsymbol u}}_{t,i}$ in ascending order. Then, we obtain ${{\boldsymbol u}}_{t,i}=[{{\boldsymbol u}}_{t,i}^{[1]}, \ldots, {{\boldsymbol u}}_{t,i}^{[j]}, \ldots, {{\boldsymbol u}}_{t,i}^{[J]}]^{\top}$ and $|{{\boldsymbol u}}_{t,i}^{[j]}|\le |{{\boldsymbol u}}_{t,i}^{[j+1]}|$. Thus, we have
\begin{align} \label{eqn:app-a-00}\small
\begin{split}
    {\mathbb E}\| {\boldsymbol u}_{t,i} \|^2 ={\mathbb E}\sum\limits_{j = 1}^J |{\boldsymbol u}_{t,i}^{[j]}|^2 = J{\mathbb E}|{\boldsymbol u}_{t,i}^{[j]}|^2=\frac{Ju_{\max}^2}{3}.
\end{split}
\end{align}
Based on Assumption \ref{asmp:5}, the update generated from local training with ${\boldsymbol w}^{\alpha}_{t,i}$ is equal to $\texttt{shrink}({\boldsymbol u}_{t,i}, \alpha_{t,i})$. The operation of model shrinking on ${{\boldsymbol u}}_{t,i}$ with $\alpha _{t,i}$ can be viewed as removing $(1-\alpha _{t,i})J$ elements with the least value from ${{\boldsymbol u}}_{t,i}$. Then, we obtain $\texttt{shrink}({\boldsymbol u}_{t,i}, \alpha_{t,i}) = [0,\ldots,0,{{\boldsymbol u}}_{t,i}^{[(1-\alpha _{t,i})J+1]},\ldots,{{\boldsymbol u}}_{t,i}^{[J]} ]^{\top}$. Thus, we have  
\begin{align} \label{eqn:app-a-1}\small
\begin{split}
    {\mathbb E}\| {\boldsymbol u}_{t,i} - \texttt{shrink}({\boldsymbol u}_{t,i}, \alpha_{t,i}) \|^2 ={\mathbb E}\sum\limits_{j = 1}^{(1-\alpha _{t,i})J} |{\boldsymbol u}_{t,i}^{[j]}|^2 \\ =J(1-\alpha _{t,i})^3u_{\max}^2/3= (1-\alpha _{t,i})^3 {\mathbb E}\|{\boldsymbol u}_{t,i}\|^2.
\end{split}
\end{align}

We next focus on the gradient compression.
The operation of gradient sparsification on ${\boldsymbol u}_{t,i}$ with sparsity of $\rho _{t,i}$ can be viewed as removing $\rho _{t,i}J$ elements with the least value from ${\boldsymbol u}_{t,i}$. Then, the quantization is conducted on the non-zero elements of $\hat{{\boldsymbol u}}_{t,i}$, and we obtain $\texttt{cmprs}({{\boldsymbol u}}_{t,i}, \beta_{t,i})=[0,\ldots,0,\tilde{{\boldsymbol u}}_{t,i}^{[\rho _{t,i}J+1]},\ldots,\tilde{{\boldsymbol u}}_{t,i}^{[J]} ]^{\top}$. Furthermore, we have
\begin{align} \label{eqn:app-a-001}\small
\begin{split}
&\;{\mathbb E} \| {{\boldsymbol u}}_{t,i} - \texttt{cmprs}({{\boldsymbol u}}_{t,i}, \beta_{t,i})\|^2\\=&\; \underbrace{{\mathbb E}\sum\limits_{j = 1}^{\rho_{t,i}J} |{\boldsymbol u}_{t,i}^{[j]}|^2}_{\text{(A)}} + \underbrace{{\mathbb E}\sum\limits_{j = \rho_{t,i}J+1}^{J} |{\boldsymbol u}_{t,i}^{[j]}-\tilde{\boldsymbol u}_{t,i}^{[j]}|^2}_{\text{(B)}}.
\end{split}
\end{align}
Likewise to Eqn. (\ref{eqn:app-a-1}), we have $\text{(A)}=\rho_{t,i}^3{\mathbb E}\|{\boldsymbol u}_{t,i}\|^2$. Based on Eqn. (\ref{eqn:quant}) and the statistical feature of ${\boldsymbol u}_{t,i}$, we obtain  $\text{(B)}=(1-\rho_{t,i})^3{\mathbb E}\|{\boldsymbol u}_{t,i}\|^2/(2L _{t,i}^2)$.

Given plain update ${\boldsymbol u}_{t,i}$ in 32-bit floating point and the desired compression rate $\beta _{t,i}$, we can set $\rho _{t,i} = 1 - \sqrt{\beta _{t,i}}$ and $L _{t,i}=2^{32\sqrt{\beta _{t,i}}}$ for the analysis. In this way, the operations of sparsification and quantization contribute equally to the gradient compression. Furthermore, we have
\begin{align} \label{eqn:app-a-002}\small
\begin{split}
{\mathbb E} \| {{\boldsymbol u}}_{t,i} - \texttt{cmprs}({{\boldsymbol u}}_{t,i}, \beta_{t,i})\|^2
\le (1-\beta_{t,i})^2{\mathbb E}\|{\boldsymbol u}_{t,i}\|^2.
\end{split}
\end{align}

Next, we focus on the local divergence $\delta _{t,i}$ with respect to $\alpha _{t,i}$ and $\beta _{t,i}$. According to the Definition \ref{defn:local-div}, we have 
\vspace{-1pt}
\begin{align}
\label{eqn:app-a-6}
\small
\begin{split}
&\;{\mathbb E}\|\delta _{t,i}\|^2 = {\mathbb E}\| {\boldsymbol u}_{t,i} - \texttt{cmprs}([{\boldsymbol u}_{t,i}]^{\alpha}, \beta_{t,i}) \|^2 \\
=  &\;{\mathbb E}\| {\boldsymbol u}_{t,i} - [{\boldsymbol u}_{t,i}]^{\alpha} \|^2 + {\mathbb E}\|[{\boldsymbol u}_{t,i}]^{\alpha} - \texttt{cmprs}([{\boldsymbol u}_{t,i}]^{\alpha}, \beta_{t,i}) \|^2\\
&\;\;+ 2\underbrace{<{\boldsymbol u}_{t,i} - [{\boldsymbol u}_{t,i}]^{\alpha}, [{\boldsymbol u}_{t,i}]^{\alpha} - \texttt{cmprs}([{\boldsymbol u}_{t,i}]^{\alpha}, \beta_{t,i})>}_{\text{(C)}}.
\vspace{-14pt}
\end{split}
\end{align}
It can be verified that the two vectors in term (C) are orthogonal, and we obtain $\text{(C)}=0$.
According to Eqns~(\ref{eqn:app-a-1}) and (\ref{eqn:app-a-002}), we further obtain
\vspace{-1pt}
\begin{align}
\small
\begin{split}
\label{eqn:app-a-7}
&\;{\mathbb E}\|\delta _{t,i}\|^2 \le (1-\alpha _{t,i})^3 {\mathbb E}\|{\boldsymbol u}_{t,i}\|^2 + (1 - \sqrt {\beta _{t,i}}  )^2{\mathbb E} \| [{\boldsymbol u}_{t,i}]^{\alpha} \|^2 \\
{\mathop  \le \limits^{(\text{a})}}&\;  (1-\alpha _{t,i})^3 {\mathbb E}\|{\boldsymbol u}_{t,i}\|^2 \\
&\;\;+ (1 - \sqrt {\beta _{t,i}}  )^2\alpha_{t,i}(\alpha_{t,i}^2-3\alpha_{t,i}+3){\mathbb E}\|{\boldsymbol u}_{t,i}\|^2\\
{\mathop  \le \limits^{(\text{b})}}&\;\big(1 - \alpha _{t,i} (2-\alpha _{t,i})\sqrt {\beta _{t,i}}  \big)^2{\mathbb E} \| {\boldsymbol u}_{t,i} \|^2.
\end{split}
\vspace{-14pt}
\end{align}
Likewise to Eqn. (\ref{eqn:app-a-1}), inequality (a) stems from the fact that ${\mathbb E} \| [{\boldsymbol u}_{t,i}]^{\alpha} \|^2=\alpha_{t,i}(\alpha_{t,i}^2-3\alpha_{t,i}+3){\mathbb E}\|{\boldsymbol u}_{t,i}\|^2$. Besides, inequality (b) holds for all $\alpha_{t,i}\in[\alpha^{\min},1]$ and $\beta_{t,i} \in [0, \beta^{\max}]$.
Thus, we complete the proof.
\end{proof*}

\section{Proof of Lemma 2}
\begin{proof}
Based on Definition \ref{defn:global-div} and Lemma \ref{lem:local-divergence}, we have
\begin{align}
\small
\begin{split}\label{eqn:app-b-1}
{\mathbb E} {\left\Vert \Delta _t  \right\Vert^2} &= {\mathbb E} {\Big\| {\sum\limits_{i = 1}^I {{p_{t,i}} } {{\boldsymbol u}}_{t,i}} -  {\sum\limits_{i = 1}^I {{p_{t,i}} } \tilde{{\boldsymbol u}}_{t,i}} \Big\|^2}\\
& \le {\mathbb E}  {\Big( {\sum\limits_{i = 1}^I {{p_{t,i}}\big(1 - \alpha _{t,i} (2-\alpha _{t,i})\sqrt {\beta _{t,i}}  \big)\left\| {\boldsymbol u}_{t,i} \right\|} } \Big)^2}.
\end{split}
\end{align}
We use $\eta$ to denote the learning rate, and ${\boldsymbol u}_{t,i}=\eta \nabla F_i({{\boldsymbol w}_{t}})$. Based on Assumption \ref{asmp:4}, we obtain
\begin{align}
\small
\begin{split}\label{eqn:app-b-2}
{\mathbb E} {\left\Vert \Delta _t  \right\Vert^2}  \le  \varepsilon \eta ^2 {\Big( {\sum\limits_{i = 1}^I {p_{t,i}}\big(1 - \alpha _{t,i} (2-\alpha _{t,i})\sqrt {\beta _{t,i}}  \big)}  \Big)^2 } {\mathbb E}  \|\nabla F({{\boldsymbol w}_t})\|^2.
\end{split}
\vspace{-14pt}
\end{align}
According to Cauchy–Schwarz inequality, we obtain
\begin{align}
\small
\begin{split}\label{eqn:app-b-3}
{\mathbb E} {\left\Vert \Delta _t  \right\Vert^2} \le I\varepsilon \eta ^2 { {\sum\limits_{i = 1}^I {p_{t,i}^2}\big(1 - \alpha _{t,i} (2-\alpha _{t,i})\sqrt {\beta _{t,i}}  \big)^2} } {\mathbb E}  \|\nabla F({{\boldsymbol w}_t})\|^2.
\end{split}
\vspace{-10pt}
\end{align}
Thus, we complete the proof.
\end{proof}

\section{On the Convergence of AnycostFL}
\begin{proof}
Inspired by the studies in \cite{chen2020joint, chen2020convergence}, we deduce the convergence analysis of AnycostFL. According to Taylor expansion and Assumption \ref{asmp:3}, we have
\begin{align}
\small
\begin{split}\label{eqn:taylor-exp}
F({{\boldsymbol w}_{t + 1}}) &\le F({{\boldsymbol w}_t}) + {({\boldsymbol w}_{t+1} -{\boldsymbol w}_{t} )^\top}\nabla F({{\boldsymbol w}_t}) + \frac{\lambda }{2}{\left\Vert {{{\boldsymbol w}_{t + 1}} - {{\boldsymbol w}_t}} \right\Vert^2} \\ 
& = F({{\boldsymbol w}_t}) - { {\tilde{{\boldsymbol u}}_t}^\top}\nabla F({{\boldsymbol w}_t}) + \frac{\lambda }{2}{\big\| {\tilde{{\boldsymbol u}}_t} \big\|^2}.
\end{split}
\end{align}
By using learning rate $\eta = \frac{1}{\lambda}$, we obtain
\begin{align}
\small
\begin{split}\label{eqn:taylor-exp-2}
{\mathbb E} \big(F({{\boldsymbol w}_{t + 1}})\big) &\le {\mathbb E} \Big( F({{\boldsymbol w}_t}) - { \lambda \left( {\boldsymbol u}_t -\Delta _t\right)^\top}{\boldsymbol u}_t + \frac{\lambda }{2}{\| {\boldsymbol u}_t -\Delta _t \|^2} \Big)\\
&= {\mathbb E} \Big(F({{\boldsymbol w}_t}) -\frac{1}{2\lambda}{\| \nabla F({{\boldsymbol w}_t}) \|^2} + \frac{\lambda}{2}{\| \Delta _t  \|^2} \Big).
\end{split}
\end{align}
We now pay attention to the upper bound of ${\left\Vert \Delta _t  \right\Vert^2}$. Based on Jensen's inequality and Eqn.~(\ref{eqn:app-b-2}), we obtain 
\begin{align}
\small
\begin{split}\label{eqn:glb-cmp-err-4}
{\mathbb E} {\left\Vert \Delta _t  \right\Vert^2} & \le    \varepsilon \eta ^2 { \underbrace{\sum\limits_{i = 1}^I {p_{t,i}}\big(1 - \alpha _{t,i} (2-\alpha _{t,i})\sqrt {\beta _{t,i}}  \big)^2}_{(\text{D})}  } {\mathbb E} \|\nabla F({{\boldsymbol w}_t})\|^2 .
\end{split}
\end{align}
By putting Eqn.~(\ref{eq:thm1}) into (A), we have
\begin{align}
\begin{split}\label{eqn:glb-cmp-err-5}
{\mathbb E} \|\text{D}\| &\le {\mathbb E} \left\|\frac{I}{{\sum\limits_{i = 1}^I {\frac{1}{{{{\left( {1 - {\alpha _{t,i}}(2 - {\alpha _{t,i}})\sqrt {{\beta _{t,i}}} } \right)}^2}}}} }}\right\|{\mathop  \le \limits^{(\text{c})}}{\mathbb E} \left\|\frac{I}{{\sum\limits_{i = 1}^I {\frac{1}{{1 - {\alpha_{t,i} ^4}\beta_{t,i} }}} }}\right\| ,
\end{split}
\end{align}
where (c) always holds for $\alpha _{t,i} \in [0, 1]$ and $\beta _{t,i} \in [0, 1]$.
According to Definition \ref{defn:learning-gain}, we have $g_{t,i}=\alpha_{t,i}^4\beta_{t,i}$ and $g_t = {\sum\nolimits_{i} {g_{t,i}}} / I$.
Since $1 / \big({\sum\nolimits_i {\frac{1}{{{1 - g_{t,i} }}}} }\big)$ is a concave function with respect to $g_{t,i}$, based on Jensen's inequality, we obtain
\begin{align}
\begin{split}\label{eqn:glb-cmp-err-6}
{\mathbb E} \|\text{A}\|  \le  {\frac{I}{{{\sum\nolimits_i {\frac{1}{{{1 - {{\mathbb E}({\alpha _{t,i}^4\beta _{t,i}})} }}}} }}}} = 1-g_{t}.
\end{split}
\end{align}
Since the training strategies of each device and the norm of the gradient of global data $\|\nabla F({{\boldsymbol w}_t})\|$ are independent, by putting Eqn.~(\ref{eqn:glb-cmp-err-6}) back to Eqn.~(\ref{eqn:glb-cmp-err-4}), we obtain
\begin{align}
\begin{split}\label{eqn:glb-cmp-err-7}
{\mathbb E} {\left\Vert \Delta _t  \right\Vert^2} & \le  {\mathbb E} \Big( \varepsilon \eta ^2 { \big(1-g_{t}\big)   }  \|\nabla F({{\boldsymbol w}_t})\|^2 \Big).
\end{split}
\end{align}

Next, by putting Eqn.~(\ref{eqn:glb-cmp-err-7}) back to Eqn.~(\ref{eqn:taylor-exp-2}), we have
\begin{align}
\small
\begin{split}\label{eqn:taylor-exp-3}
{\mathbb E} \big(F({{\boldsymbol w}_{t + 1}})\big) &\le   {\mathbb E} \Big(F({{\boldsymbol w}_t}) -\frac{1+\varepsilon\big(g_t-1\big)}{2\lambda}{\| \nabla F({{\boldsymbol w}_t}) \|^2}\Big).
\end{split}
\end{align}
Subtracting $F({\boldsymbol w}^\ast)$ in both sides of Eqn.~(\ref{eqn:taylor-exp-3}) yields
\begin{align}
\small
\begin{split}
\label{eqn:taylor-exp-4}
{\mathbb E} &\big(F({{\boldsymbol w}_{t + 1}} - F({\boldsymbol w}^\ast)\big) \\
&\le {\mathbb E} \Big(F({{\boldsymbol w}_t}) - \frac{1+\varepsilon(g_t-1)}{2\lambda}{\| \nabla F({{\boldsymbol w}_t}) \|^2} - F({\boldsymbol w}^\ast)\Big).
\end{split}
\end{align}
Based on Assumptions \ref{asmp:2} and \ref{asmp:3}, we have \cite{chen2020joint, boyd2004convex}
\begin{align}
\begin{split}\label{eqn:cvx}
\|\nabla F({{\boldsymbol w}_t})\|^2 \ge 2\nu\big(F({{\boldsymbol w}_t}) - F({{\boldsymbol w}^\ast}) \big).
\end{split}
\end{align}
Plugging Eqn.~(\ref{eqn:cvx}) into Eqn.~(\ref{eqn:taylor-exp-4}), we have
\begin{align}
\begin{split}\label{eqn:taylor-exp-5}
{\mathbb E} \big(F({{\boldsymbol w}_{t + 1}}) - F({\boldsymbol w}^\ast)\big) &\le Z_t{\mathbb E}  \big(F({{\boldsymbol w}_{t}}) - F({\boldsymbol w}^\ast)\big),
\end{split}
\end{align}
where $Z_t=1 - \frac{\nu }{\lambda }\left( {1 - {\varepsilon}( {1 - g_t } )} \right)$.

Let $g^{\min} = \mathop {\min } \{g_t\}_{\forall t}$ be the minimal global learning gain over $T$ global rounds. By recursively applying the above inequality from iteration round 0 to $T$, we can obtain 
\begin{align}
\begin{split}\label{eqn:thm-2}
{\mathbb E} \big(F({{\boldsymbol w}_{T}}) - F({\boldsymbol w}^\ast)\big) &\le Z^{T-1} {\mathbb E} \big(F({{\boldsymbol w}_{0}}) - F({\boldsymbol w}^\ast)\big),
\end{split}
\end{align}
where $Z=1 - \frac{\nu }{\lambda }\left( {1 - {\varepsilon}( {1 - g^{\min}} )} \right)$. Thus, we complete the proof.
\end{proof}

\bibliographystyle{IEEEtran}
\bibliography{reference.bib}

\end{document}